\documentclass{ifacconf}

\makeatletter
\let\old@ssect\@ssect 
\makeatother
\usepackage{optidef}
\usepackage{graphics} 
\usepackage{times} 
\usepackage{amsmath} 
\usepackage{amssymb}  
\usepackage[hidelinks]{hyperref}
\usepackage[printonlyused]{acronym}
\usepackage{nomencl}
\usepackage{mathrsfs}
\usepackage[caption=false,font=footnotesize]{subfig}
\newcommand{\norm}[1]{\left\lVert#1\right\rVert}
\usepackage{makecell}
\usepackage{longtable}
\usepackage{graphicx}      
\usepackage{natbib}        

\makeatletter
\def\@ssect#1#2#3#4#5#6{%
  \NR@gettitle{#6}
  \old@ssect{#1}{#2}{#3}{#4}{#5}{#6}
}
\makeatother

\begin{document}
\begin{frontmatter}

\title{Continuous-Time Trajectory Optimization for Decentralized Multi-Robot Navigation}
%
%
%

\author{Shravan Krishnan,}
\author{Govind Aadithya Rajagopalan,} 
\author{Sivanathan Kandhasamy}
\author{and Madhavan Shanmugavel}
\address{Autonomous Systems Lab, Department of Mechatronics, SRM Institute of Science and Technology, India, e-mail: shravan\_krishnan@srmuniv.edu.in }
   



\begin{abstract}
Multi-robot systems have begun to permeate into a variety of different fields, but collision-free navigation in a decentralized manner is still an arduous task. Typically, the navigation of high speed multi-robot systems demands replanning of trajectories to avoid collisions with one another. This paper presents an online replanning algorithm for trajectory optimization in labeled multi-robot scenarios. With reliable communication of states among robots, each robot predicts a smooth continuous-time trajectory for every other remaining robots. Based on the knowledge of these predicted trajectories, each robot then plans a collision-free trajectory for itself. The collision-free trajectory optimization problem is cast as a non linear program (NLP) by exploiting polynomial based trajectory generation. The algorithm was tested in simulations on Gazebo with aerial robots. 
\end{abstract}


\end{frontmatter}

\section{Introduction}
Multi-robotics is currently an emerging field with numerous applications and potential in several cross-cutting fields. However, the navigation of a group of robots in complex environments is still a daunting task. In such scenarios, it is essential that every robot is able to autonomously plan collision-free trajectory while also making sure that it replans the generated trajectory appropriately based on the dynamic nature of environment. In this paper, We present a decentralized multi-robot trajectory optimization algorithm which accepts only current states and non-interchangeable end poses of other robots as inputs and generates smooth trajectories. The algorithm exploits the differential flatness property of the robots  \citep{mellinger2011minimum} \citep{chul2010}; a property  that allows planning of trajectories in the space of flat variables and their derivatives.  

Trajectory generation for a fleet of robots has received wide spread attention recently. A plethora of different solutions have been proposed to solve this problem. However, a majority of them are centralized methods \citep{tang2018hold}\citep{Solovey2016finding}. Centralized methods have recently branched into decentralized methods that are capable of planning trajectories appropriately in a decentralized manner \citep{Bekris2017Safe} \citep{2017fast}. These approaches with exception of \citep{tang2018hold}(which is also centralized) are discrete-time and provide the robots with discrete commands and thereby overlook probable collisions. Moreover, discrete approaches are sampling based \citep{Solovey2016finding}\citep{Bekris2017Safe} or search based \citep{fan} that incrementally look through the space. 

An another class of multi-robot algorithms  utilize the concept of velocity obstacles to plan velocities appropriately by formulating admissible velocities that robots can be at without colliding \citep{Berg2008,mora2018cooperative,Jur2011,rufli2013reciprocal,snape2010smooth}. The advantage of such approaches is that they are easily adaptable for decentralized implementations, but restricted to only circular shaped robots. Recently, some extensions have been proposed  that allow their usages for heterogeneous robots \citep{bareiss2017general}. Building on these works, \citep{mora2018cooperative} proposed a collaborative collision avoidance for non-holonomic robots with re-planning, while respecting the vehicular constraints and also accounting for potential tracking error bounds of the robot. They have also proposed an extension of the same to aerial robots in \citep{aerialalonsomora}. 

\begin{figure*}
\centering
\subfloat[][]{\includegraphics[width=0.35\textwidth]{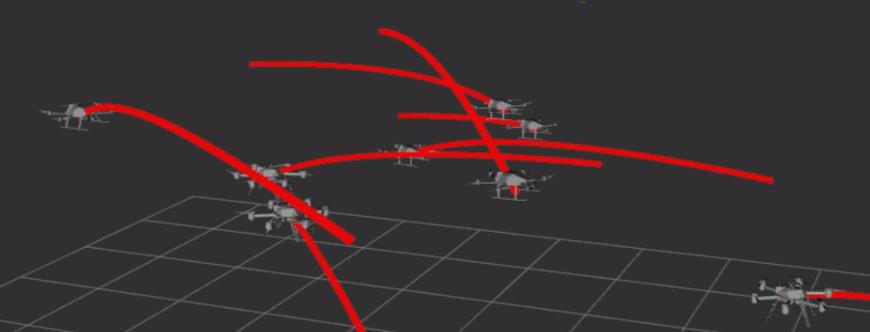}\label{maneuver/maneuver1}} 
\hspace{2mm} \subfloat[][]{\includegraphics[width=0.25\textwidth]{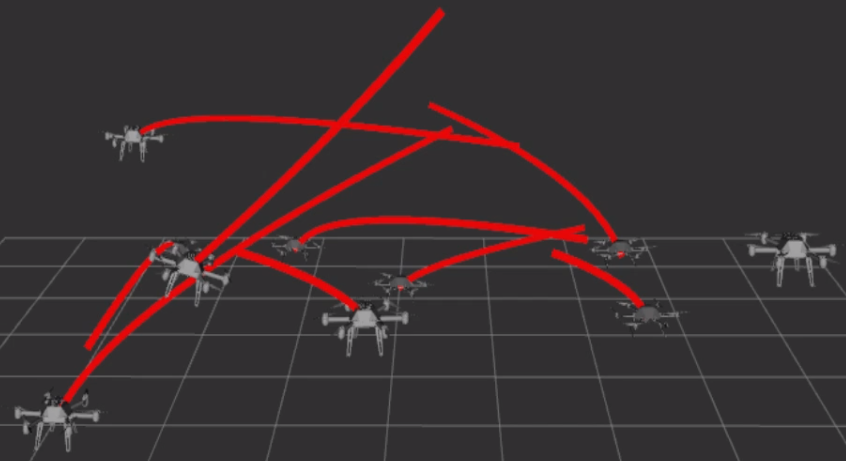}\label{maneuver2}} 
\hspace{2mm} \subfloat[][]{\includegraphics[width=0.3\textwidth]{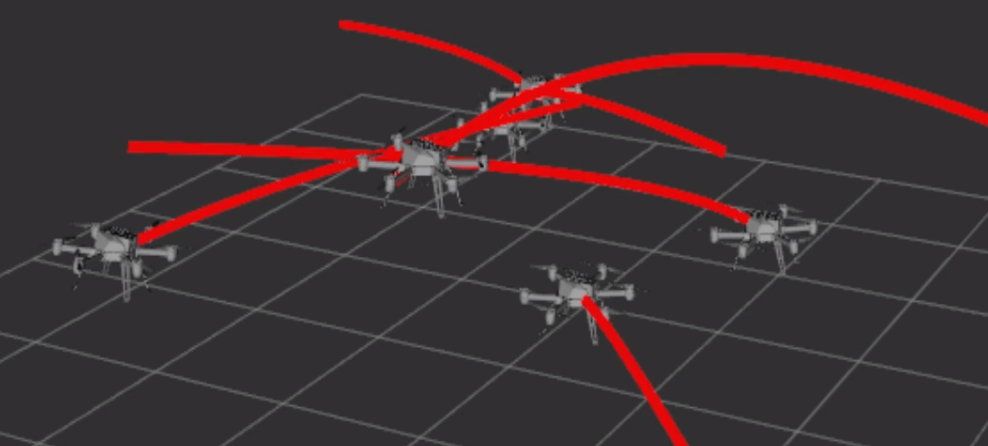}\label{maneuver3}} 
\caption{A sequence of images showing robots during different transitions. The red curves are the planned trajectories. A video of the simulations is available at \href{https://bit.ly/2ZpGq76}{https://bit.ly/2ZpGq76}}
\label{differentposes}
\end{figure*}

Recently, a centralized method for generating collision-free trajectories for a swarm of quadrotors in known obstacle filled environments was proposed \citep{honig2018}. The method utilized a three step process with first step for generating sparse roadmaps,  second step for planning discrete schedules and third step for generating bezier curve based time parametrized smooth trajectories. Additionally, the duration of the trajectory is not directly optimized but scaled to satisfy the dynamics of robots. 


Methods based upon sequential convex programming have also been proposed for multi-robot systems \citep{chen2015scp}\citep{agualiro2012scp}. However, these methods are centralized and use a discrete approach for planning trajectories. Furthermore, they use only second order dynamics with jerk constrained as rate of change between two subsequent time steps. Nevertheless, the algorithms do not have any dimensional restrictions. Additionally, utilizing such approaches is tractable only if the end times are known or end times are scaled accordingly beforehand.

Barrier functions based methods have been proposed for collision avoidance \citep{safe2017certificate}. Recently, model predictive control based methods have been used for multi-robot collision avoidance \citep{scpmpc}\citep{rob}. \citep{mpcorca}  proposed a Model Predictive Control scheme based on \citep{Jur2011}.

A majority of these methods account for lower order dynamics of the robot and/or do not guarantee a continuous time trajectory that also accounts for collision avoidance in continuous time. Moreover, the algorithms do not consider time as a coupled optimization parameter. 

In this work, a higher order dynamics(three) is used for the robots rather than two which is often found in literature. Additionally, trajectories are generated in continuous-time for multi-robot systems with $M$ different robots rather than sampling based or discrete time methods.

The major contributions of this paper are:
\begin{enumerate}
\item A closed form solution for generating minimum time-jerk squared smooth trajectories given current state and desired end position.
\item A decentralized algorithm for generating collision-free continuous-time trajectories for multi-robot systems in $N$ dimensions.
\item Extensive simulations of the proposed algorithm using a variety of different aerial robots in three dimensional spaces
\end{enumerate}

 The proposed algorithm involves $MN$ polynomials in each of the $M$ robots. $(M-1)N$ polynomials denote the predicted trajectories of other robots in $N$ dimensions and $N$ polynomials represent the collision-free trajectory. 
 
 Based upon the $MN$ polynomials, a two-step process is used to generate collision-free trajectories. The first step generates the $(M-1)N$ polynomials representing the continuous-time predicted trajectory of other robots in the environment. The second step formulates a non linear optimization problem(NLP) with objectives of minimizing jerk and time while not exceeding the dynamic limits and avoiding collision(with respect to the trajectories from the previous step). The solution of the NLP provides the coefficients of the collision-free trajectory and the duration of the trajectory.


Furthermore, replanning of trajectories is done online for two reasons; one, the number of robots may not be known beforehand and two, the trajectory prediction does not account for robot-robot interaction and therefore the predicted trajectory diverges from the planned trajectories of other robots considerably during longer durations. To allow for efficient use of the previous trajectory, the algorithm is implemented in a receding horizon manner, i.e, a part of the trajectory is applied after which the trajectory is re-planned.

The rest of the paper is organized as: Section \ref{prediction} showcases the Trajectory prediction method. Section \ref{traj optimization} details the trajectory optimization problem. Section \ref{results} showcases the performance of the algorithm in simulations and Section \ref{conclusion} concludes the paper

\section{Trajectory Prediction}
\label{prediction}
Knowing the trajectories of other robots in the environment improves the efficiency of the generated trajectory in avoiding collision. In this section, we derive a closed-form solution for efficiently predicting minimum time and jerk smooth trajectories for other robots in the environment. The dynamic limits of other robots are unknown and hence neglected. The trajectory prediction algorithm was inspired by the closed form solutions for efficient trajectory generation for quadcopters shown by \citep{efficient2015quad} and is discussed below.	

Commonly used dynamic models for mobile robots or multirotors are differentially flat. This allows us to formulate the input of those robots as the $n$\textsuperscript{th} derivative of a $n$\textsuperscript{th} order robot. Additionally, The generated trajectory should be represented by at least a $2 n-1$ order polynomial \citep{tang2018hold}.

\subsection{State model}
The robots are modelled as a third order system with state $\textbf{x}=[p \hspace{1mm} v \hspace{1mm} a]$ and input $u=[j]$, where $p$ is position, $v$ is velocity, $a$ is acceleration and $j$ is jerk. Hence, the dynamics is represented as:
\begin{equation}
\dot{\textbf{x}}=[v \hspace{1mm} a \hspace{1mm} j]
\label{dynam}
\end{equation}

\begin{figure*}
    \centering
    \includegraphics[width=0.7\textwidth]{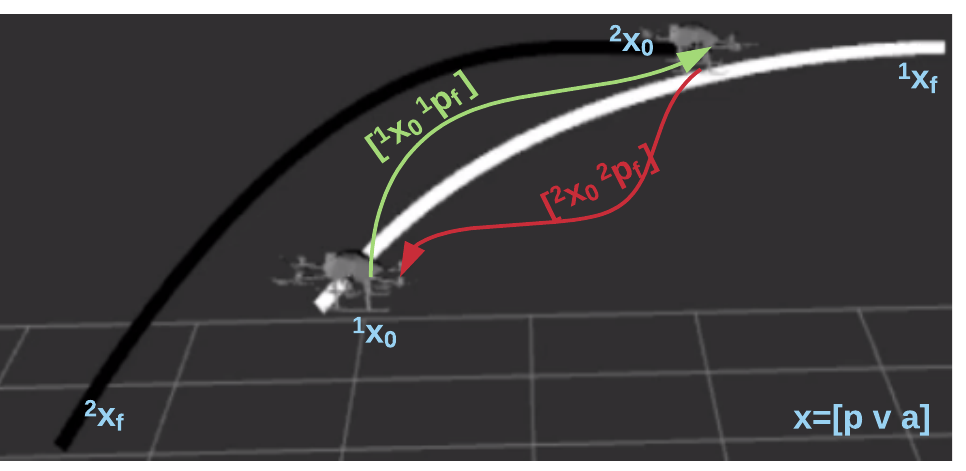}
    \caption{A schematic explaining the overall system. The white curve indicates the planned trajectory by robot 1 from it's current state $^1x_0$ to it's end state $^1x_f$. Similarly, the black curve indicates the planned trajectory by robot 2 from it's current state $^2x_0$ to it's end state $^2x_f$. The green and red arrows indicate the communication and the data being shared during the communication}
    \label{schematic}
\end{figure*}

\subsection{Objective Function}
The objective of the trajectory prediction is to find smooth and minimum-time trajectory that moves the robot from it's current state to the partially defined end state in  minimal time. The objective can be represented as 
\begin{argmini}
{}{\int_{0}^{T}  \norm{u} ^2 +1 \hspace{1mm} dt}{}{}
\end{argmini}

The objective is coupled only by the end time $T$ in each dimension. This allows looking at each dimension individually. Hence, for the sake of brevity and readability, the trajectory is derived for a single dimension and the dimensions are coupled at end once a closed-form solution for $T$ is found. 

The system represented in Eq.\eqref{dynam} is a linear system. Utilizing Pontygarin's maximum principle \citep{dynamicprogramming}, the system can be solved. We derive the closed form solution for the system with a third order model, which has been used for a variety of classes of robots ranging from multirotors \citep{efficient2015quad} to Autonomous ground vehicles \citep{highspeed}.
The Hamiltonian of the third order system is:

\begin{equation}
H(x,\lambda,u)= \norm{u}^2 +1 + \lambda_1v + \lambda_2a + \lambda_3u
\end{equation}


where $\Lambda=[\lambda_1 \hspace{2mm} \lambda_2 \hspace{2mm} \lambda_3]^T$ is the costate equation. The optimal costate equation can be formulated as:
\begin{equation}
\dot{\Lambda}^*(t)=-\frac{\partial H}{\partial x}=-\begin{bmatrix}
0 \\
\lambda_1 \\
\lambda_2 
\end{bmatrix}
\end{equation}

The solution of costate differential equations is then represented as:
\begin{equation}
\Lambda^*(t)=-\begin{bmatrix}
2\beta_1 \\
2\beta_1t+ 2\beta_2 \\
\beta_1t^2+ 2\beta_2t + 2\beta_3
\end{bmatrix}
\label{costatesol}
\end{equation}

The optimal control input can be found by:
\begin{argmini}
{u^*}{H(x^*(t),u^*(t),\Lambda^*(t))}{}{}
\end{argmini}
\begin{equation*}
u^*=-\frac{\lambda_3}{2}=\frac{\beta_1t^2}{2}+ \beta_2t + \beta_3
\end{equation*}

On integrating

\begin{equation}
\begin{split}
p^*(t)&= \frac{\beta_1t^5}{120}+\frac{\beta_2t^4}{24}+\frac{\beta_3t^3}{6}+\frac{a_0t^2}{2}+v_0t+p_0 \\
v^*(t)&=\frac{\beta_1t^4}{24}+\frac{\beta_2t^3}{6}+\frac{\beta_3t^2}{2}+a_0t+v_0 \\
a^*(t)&=\frac{\beta_1t^3}{6}+\frac{\beta_2t^2}{2}+\beta_3t+a_0
\end{split}
\label{optimal_state}
\end{equation}

As the end state is partially defined (only position), the position can be substituted into Eq. \eqref{optimal_state} resulting in 
\begin{equation}
p_{end}= \frac{\beta_1T^5}{120}+\frac{\beta_2T^4}{24}+\frac{\beta_3T^3}{6}+\frac{a_0T^2}{2}+v_0T+p_0 
\end{equation}

 The unknown coefficients can be found as the corresponding costates will be zero at the free endstates\citep{dynamicprogramming}. Thus, from \eqref{costatesol}, the  appropriate equations(second and third) can be used to solve for the three unknown coefficients. This also allows a representation of the unknown coefficients as a function of end time, known initial states and end position. Resulting in:

\begin{equation}
\begin{bmatrix}
\beta_1 \\
\beta_2 \\
\beta_3 \\
\end{bmatrix} = \frac{1}{T^5}\begin{bmatrix}
20 \\
-20T \\
10T^2
\end{bmatrix}(p_{end}-(p_0+v_0T+ \frac{a_0T^2}{a}))
\label{equation}
\end{equation}
Furthermore, due to the time being a variable to optimize, $H(x,\lambda,u)$ of the system is now zero instead of a constant \citep{dynamicprogramming}. Hence,

\begin{equation}
H(x,\lambda,u)= \norm{u}^2 +1 + \lambda_1v + \lambda_1a + \lambda_3u=0
\label{Hamil}
\end{equation}

substituting Eq. \eqref{equation} into Eq. \eqref{Hamil} and simplifying results yields

\begin{equation}
\begin{split}
&\frac{a_0^2T^8}{2}+a_0v_0T^7+(1+a_0(p_{end}-p_0))T^6+(20a_0^2v_0)T^5\\
&+(40a_0v_0+10a_0^2-717a_0^2/4)T^4+40a_0(p_{end}-p_0)T^3\\
&-697a_0v_0T^3+(20a_0(p_{end}-p_0)+20v_0+10a_0v_0)T^2\\
&-717(v_0+p_{end}a_0-a_0p_0)T^2-(1434v_0(p_{end}-p_0)T\\
&-717(p_0^2+p_{end}^2-2p_{end}p_0)
\end{split}
\end{equation}
Similarly, the other dimensions can also be lumped into the coefficients for finding the time. 
This eighth order polynomial can be solved for the roots. This is solved using \citep [p. ~ 146--147]{horn1985cr}, which is implemented in numpy on python\footnote{\href{https://docs.scipy.org/doc/numpy/reference/}{https://docs.scipy.org/doc/numpy/reference/}}. 
The trajectory with the least cost among all the real and positive solutions for the polynomial is utilized after the coefficients are found using Eq. \eqref{equation}. There are scenarios where no real roots exist,then we assume a constant jerk and use Newton's second equation of motion to solve for the end time(This assumption is used rather than fixing an end time due to the fact that fixing an end time beforehand reduces a degree of freedom in the solution). After getting the end time, we solve for the polynomial coefficients. Each of the polynomial represents the predicted trajectory in one dimension for one robot. Therefore, for every other robot that transmits it's current state and desired state, the above method is used for predicting trajectories thereby generating a smooth trajectory representation for dynamic obstacles.

\section{Trajectory generation}
\label{traj optimization}
The trajectory generation problem can be formulated as:
\begin{argmini!}
  {\textbf{\textit{x}}}{\int_{0}^{T} C_{d} +\sum_{i=1}^M C_{c}+C_{l}+ K_t \hspace{1mm} dt  }{}{}
  \label{traj_objective}
  \addConstraint{\textbf{\textit{x}}(0)}{=x_0}{}
  \label{staring_const}
  \addConstraint{\textbf{\textit{x}}(T)}{=x_f}
  \label{end_const}
\end{argmini!}

Where $C_{d}$,$C_{c}$ and $C_{l}$ are costs for trajectory smoothness, collision with other robots and dynamic limits respectively. Please note the addition of $K_t$ at the end of the cost, which is added to minimize the time taken along the trajectory. Besides, the constraint in Eq. \eqref{end_const} can be defined for the complete state or partial state. Furthermore, we represent the trajectories by time parameterized polynomials of order five to allow trajectories to be similar in representation to the predicted trajectory. 
Therefore, for each dimension, the trajectory can be represented by:
\begin{equation}
x(t)= \sum_{j=0}^{5}\alpha_j t^j
\label{decision with degree}
\end{equation}
 The decision variables of the optimization problem are:
 \begin{equation}
\mathcal{D} = [\alpha_0 \hspace{1mm} \alpha_1 \hspace{1mm} \alpha;_2 \hspace{1mm} \cdots \alpha_{6N-1} \hspace{1mm} T]^T
\label{per_vehicle}
\end{equation}

where the first six numbers represent the polynomial coefficients of the first dimension, the second six the second dimension. Lastly the $T$ represents the end time.




\subsection{Dynamic Smoothness}
To ensure that the generated trajectory is smooth, a smoothness constraint is added. This constraint is represented as:
\begin{equation}
C_{d}=Q_{dynm}\norm{\frac{d^{n}\textbf{\textit{x}}}{dt^{n}}}^2 
\label{derivative}
\end{equation}
$Q_{dynm}$ is a weight for the dynamics. This derivative cost can be integrated analytically and solved for in terms of the optimization variables as shown in Eq. \eqref{per_vehicle}.

\subsection{Collision Avoidance}
To avoid collisions with other robots in the environment, an exponential barrier function based collision avoidance method is utilized. This function can be represented as


\begin{equation}
C_{c}=Q_{obs}c(x(t))
\end{equation}
Where $Q_{obs}$ is collision avoidance weighing parameter, 
\begin{equation}
c(x(t)) = \frac{x(t)-x_{obs}(t)(v(t)-v_{obs}(t))}{\exp^{K_p(d(x(t),x_{obs}(t))-\rho)}d(x(t),x_{obs}(t))} 
\label{Pieceswise obstacle}
\end{equation}
$c(x(t))$is the collision avoidance cost and $d(x)$ is the euclidean distance between the robot and obstacle and $\rho$ is a threshold distance beyond which the robots have to be for collision-free navigation. Thus, we model the robots as $N$ dimensional spheres and $\rho$ is the sum of the two radii(minimum distance to ensure collision avoidance). The robots are modeled as spheres as it is easier to compute distances for spheres than ellipses. 

\begin{figure}
\includegraphics[width=0.5\textwidth]{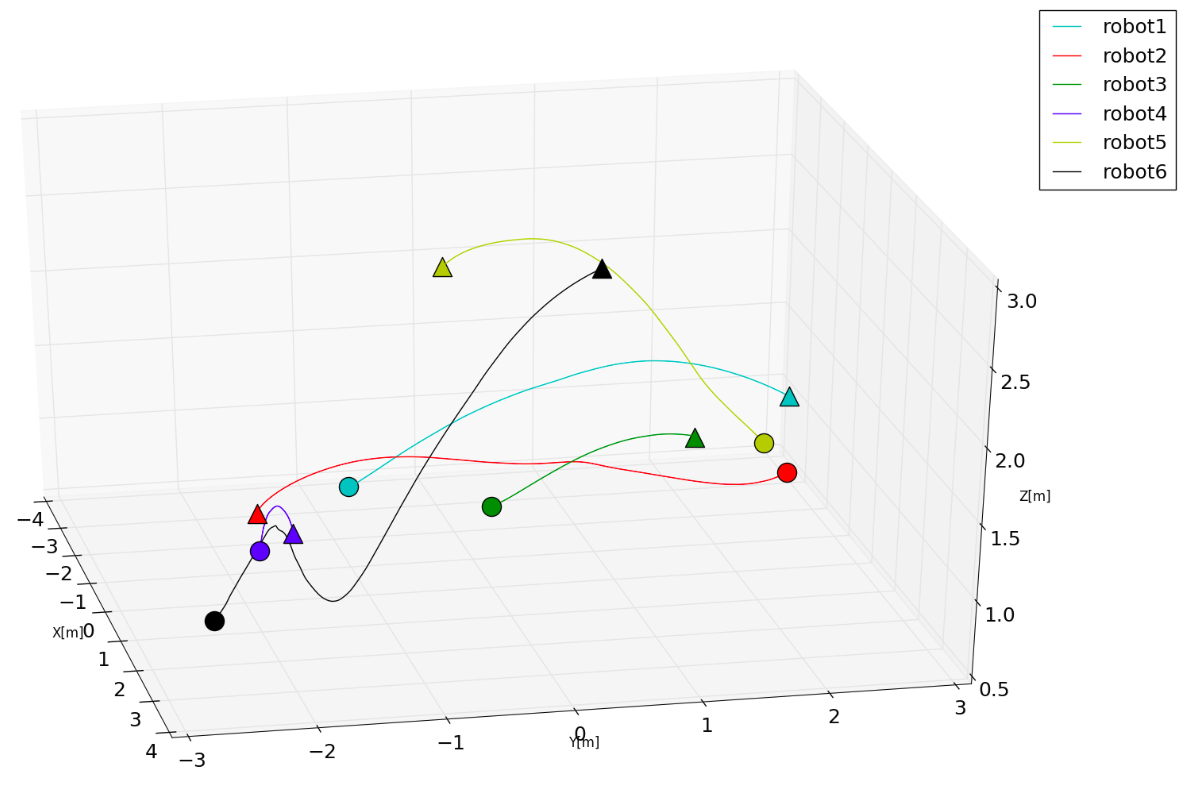}
\caption{The trajectories for six Neos robots during a rest-to-rest experiment.}
\label{six_robot}
\end{figure}


Furthermore, the velocity difference $v(t)-v_{obs}(t)$ is added to the cost for two reasons, one, to account for the difference in velocity of the robots and two, analytical integration of the proposed cost for closed-form solutions. The analytically integrated cost is then utilized for collision avoidance. Moreover, in the case of other robots( whose trajectories are predicted as detailed in Section \ref{prediction}), the trajectories are represented by time parameterized polynomials. Therefore, this time parameterized trajectories of each of the robots can be directly incorporated into the objective in $x_{obs}(t)$,  allowing for continuous time evaluation.

\subsection{Dynamic limits}
As the algorithm attempts to minimize the time taken by the robots, it is important to consider the dynamic limits of the robot. To account for the limits, we utilize a soft constraint inspired by \citep{2017real} that allows for continuous time limit verification while also not adding constraints. The dynamic limits are represented as:
\begin{equation}
C_{l}=Q_{lim}D(x(t))
\end{equation}
Where $Q_{lim}$ is the tuning weight for the dynamic limits,
\begin{equation}
D(x(t))=\exp^{d(x(t))^2-\tau^2}
\end{equation}
Here $\tau$ is the maximum limit of the specific derivative that the robot is allowed and $d(x(t))$ is the euclidean norm of the $N$ dimensions of the robot.
Furthermore, in our implementation we utilize the euclidean norm of the derivatives to account for dynamic limits rather than each dimension individually. This is done as the dynamic limits of most robots are better represented by magnitude rather than decoupled limits on each dimension. We penalize the limits for the dynamics from the first derivative to $n^{\text{th}}$ derivative of the robot's position. The usage of soft constraints is guided by it allowing for a continuous time dynamic limit checking and if necessary, the dynamic limits can be violated by a small margin by the algorithm.  

\begin{figure}
\includegraphics[width=0.5\textwidth]{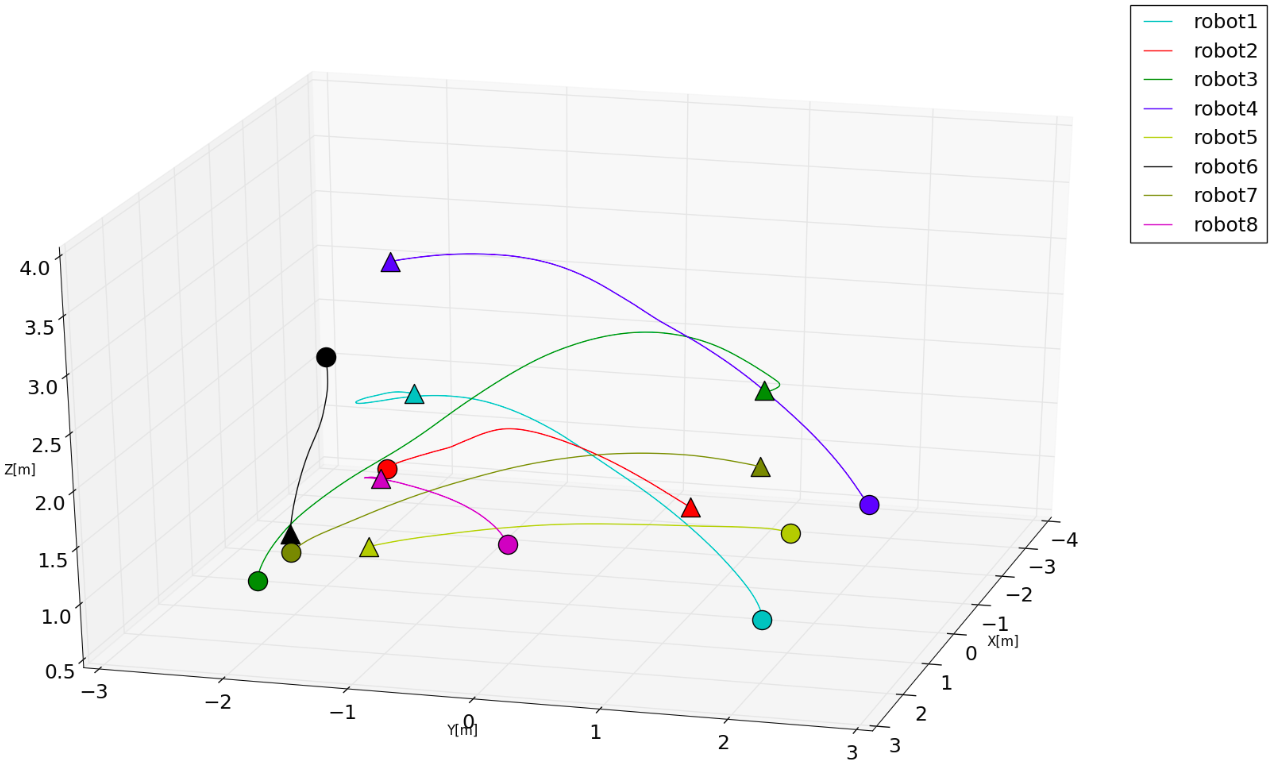}
\caption{The executed trajectory for Eight Fireflys from a hovering start to non zero end velocity. The robots as they start in a cluttered environment showcase some drastic changes and evasions, but as time progresses, they come to stand still}
\label{8neo}
\end{figure}

\begin{figure}
\includegraphics[width=0.45\textwidth]{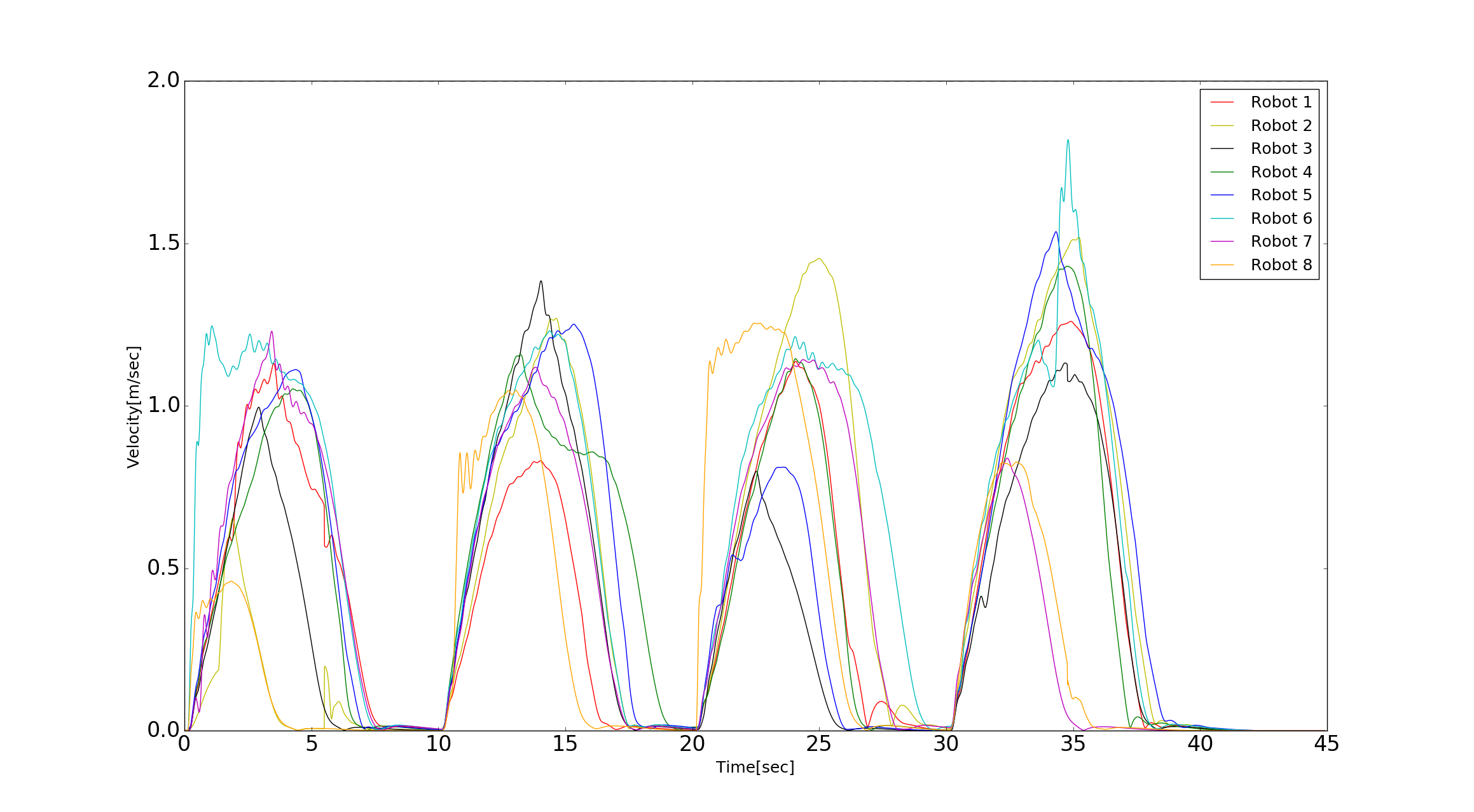}
\caption{The velocity profile for the eight robot multiple transitions maneuver. The evasive velocity profiles as the robots navigate between the other robots initially. As the robots move further away from each other, the robots velocity profiles stabilize into a smooth curve. The dynamic limits are shown in dotted black lines }
\label{eightneo}
\end{figure}



The NLP in Eq.(12) cannot be proven to be convex due to the equality constraints in Eq. \eqref{staring_const} and Eq. \eqref{end_const}. Hence, any solution that is generated is going to be locally optimal in general. We utilize Sequential Quadratic Program(SQP) to solve the NLP. The algorithm was implemented in \citep{nlopt} using the method proposed in \citep{sqp}. The initialization of the SQP is done using the previously optimized trajectory. For solving the problem during the first iteration, the algorithm was initialized using the method for trajectory prediction in Section \ref{prediction}

\begin{figure}
\centering
\includegraphics[width=0.5\textwidth]{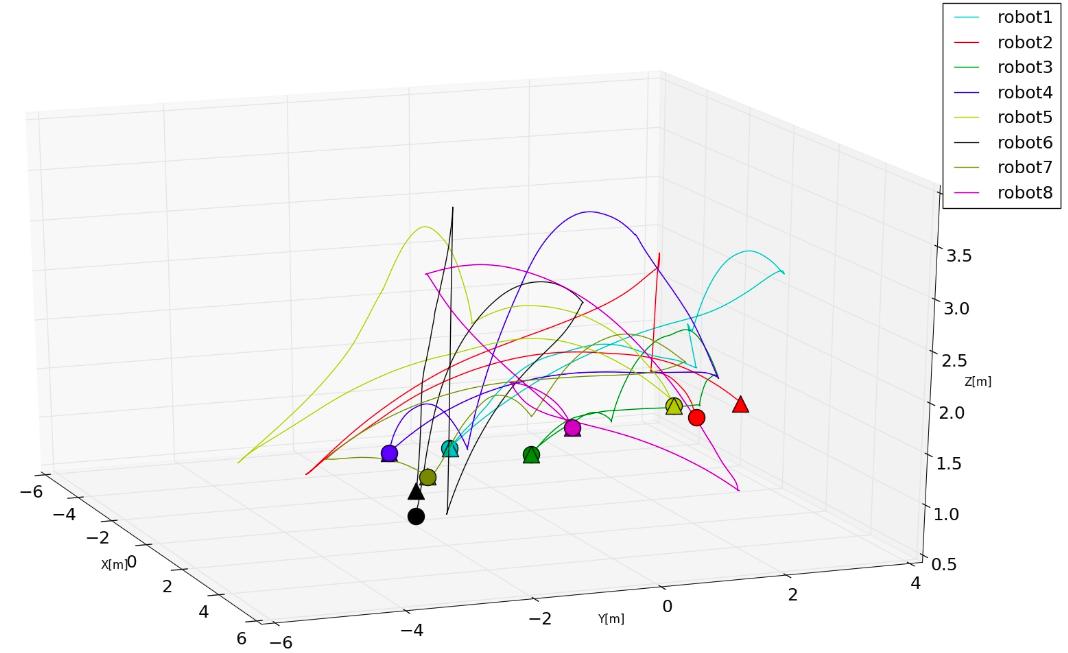}
\caption{The trajectories for Eight aerial robots(Three Neos and Five Fireflys) during multiple tranisitions}
\label{eight_robot_traj}
\end{figure}


\section{Simulation Results}
\label{results}
The algorithm was implemented in Python using Robot Operating System(ROS) with $N=3$. The algorithm was tested on a workstation with Intel Xeon E5 1630v5  processor, 64GB of RAM and a Nvidia Quadro M4000 GPU. The algorithm was verified with different rotary winged flying robots of different sizes (Asctec Hummingbird\footnote{\href{http://www.asctec.de/uav-uas-drohnen-flugsysteme/asctec-hummingbird/}{http://www.asctec.de/uav-uas-drohnen-flugsysteme/asctec-hummingbird/}},Firefly\footnote{\href{http://www.asctec.de/uav-uas-drohnen-flugsysteme/asctec-firefly/}{http://www.asctec.de/uav-uas-drohnen-flugsysteme/asctec-firefly/}},Neo\footnote{\href{http://www.asctec.de/uav-uas-drohnen-flugsysteme/asctec-neo/}{http://www.asctec.de/uav-uas-drohnen-flugsysteme/asctec-neo/}}) in Gazebo using RotorS \citep{rotors}, a high fidelity multirotor simulator.

\begin{figure}
\includegraphics[width=0.5\textwidth]{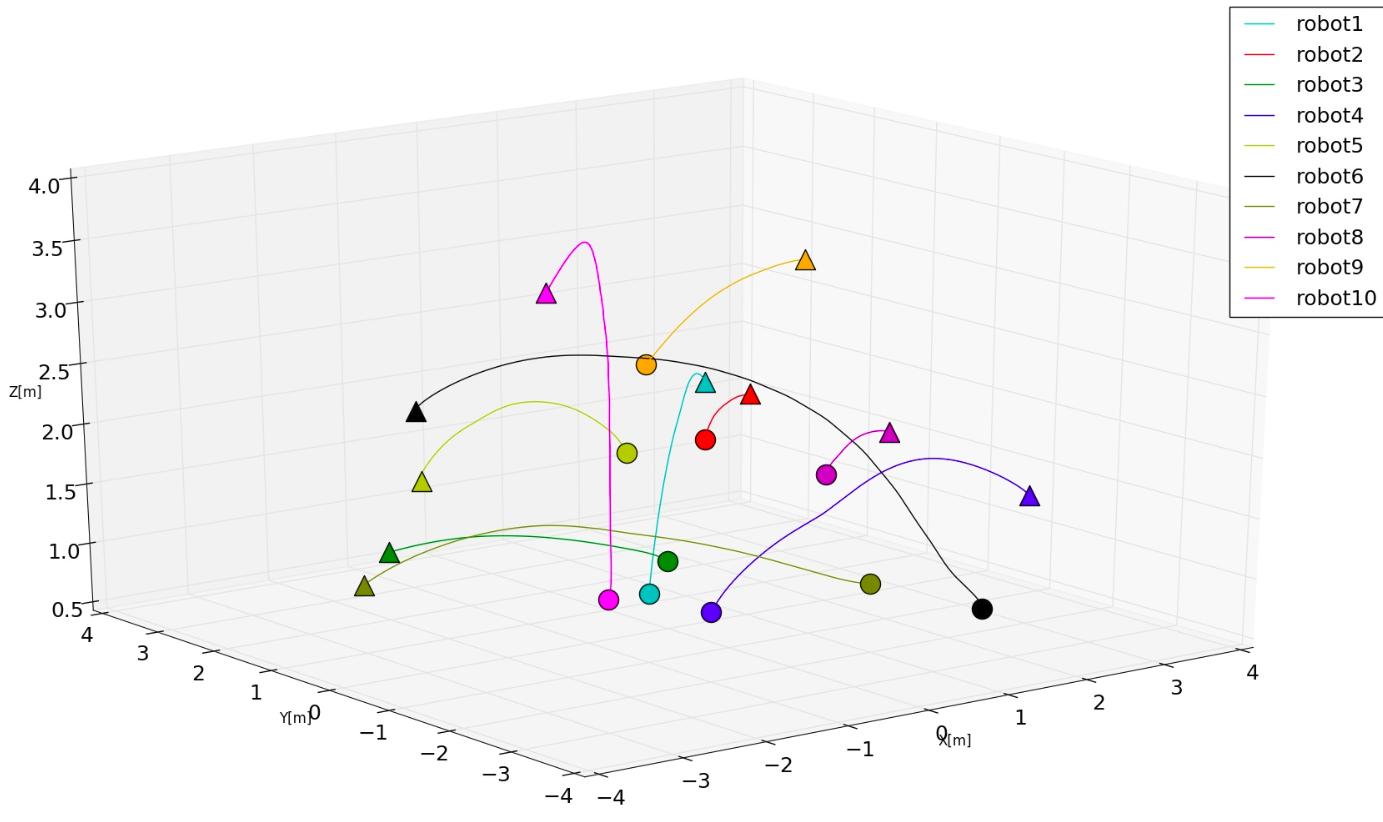}
\caption{Five Fireflys and Five Neos's trajectories during a rest-to-rest maneuver}
\label{10hetero}
\end{figure}

Although aerial robots are fourth order systems, in our experiments we utilize third order system that minimizes the jerk of the robot similar to \citep{efficient2015quad}. To track the generated trajectory, we use \citep{taecontroller}. Throughout the experiments, the yaw of the system is kept free as the rotation about yaw does not affect the translational motion. The trajectories are re-planned at a frequency of 8Hz.     
\subsection{Simulations}
In Figure \ref{six_robot},\ref{8neo},\ref{eight_robot_traj} and \ref{10hetero}, the colored circles represent the starting point of the robots while the colored triangles represent the end points of the robots.

In a first simulation, six Neos were spawned randomly and tested for rest-to-free-end-state trajectory. 
The plotted trajectories are shown in Figure \ref{six_robot}. 

In a second simulation, eight Fireflys starting from a tightly scrunched space, with many robots stacked on top of each other were given a non-rest end state. This experiment also had the robots start off at distances closer than the allowed threshold. The trajectories of the robots during this experiment are shown in Figure \ref{8neo}. 

In a third simulation, the algorithm's performance with heterogeneous robots and multiple maneuvers was tested (The robots were first given a rest-to-rest state, and after that another non-rest desired pose was published). The experiment used five Fireflys and three Neos and consisted of four different transitions with the latter three transitions having non-rest end states (left free). The robots were able to maneuver safely in the experiment while ensuring the dynamic limits were not exceeded. The resulting trajectory is shown in Figure \ref{eight_robot_traj}. 

It can also be seen from the velocity profile that some of the robots do come to rest during the third transition.

In a final simulation, five Fireflys and five Neos were tested for their capabilities with rest-to-rest transitional maneuver. The resulting trajectories for different robots are shown in Figure \ref{10hetero}.

The algorithm is capable of generating collision-free smooth trajectories for the robots to traverse across the environment. Furthermore, the utilization of soft constraints allows the robots to be able to violate the constraints if required. This is while being a boon for dynamic limits but is not so for collision avoidance.

\section{Conclusion}
\label{conclusion}
A decentralized algorithm for replanning trajectories for multi-robots systems with third order dynamics was proposed in this paper. Each robot uses the algorithm to predict continuous-time trajectories for other robots in the environment and utilizes those trajectories for planning collision-free trajectory for itself. The algorithm was simulated extensively on Gazebo for up to ten robots with speeds of up to 2 m/s in three dimensional spaces. Furthermore, the algorithm utilizes continuous time trajectory parameterization and implements soft constraints for dynamic limits.

Future work is to implement the algorithm on real robotic platforms. For collision avoidance, the performance is dependent on the axis in which the robots are proximal to each other. This is an anomaly that we look to solve in the future. Another avenue for research is to develop a methodology that allows for adapting the weights online as hand tuning the cost function is tedious and time consuming. Besides, in most of the experiments, the algorithm predicts trajectories assuming non-zero end states, developing a method to decide whether to use free-end state or rest-end state is another point to look through in the future.  Moreover, accounting for deviations in the predicted trajectory and actual trajectory can be incorporated into the collision avoidance to allow for better predictions.


\bibliography{references,refer}

\begin{thebibliography}{4}
\providecommand{\natexlab}[1]{#1}
\providecommand{\url}[1]{\texttt{#1}}
\providecommand{\urlprefix}{URL }
\expandafter\ifx\csname urlstyle\endcsname\relax
  \providecommand{\doi}[1]{doi:\discretionary{}{}{}#1}\else
  \providecommand{\doi}{doi:\discretionary{}{}{}\begingroup
  \urlstyle{rm}\Url}\fi

\bibitem[{Able(1956)}]{Abl:56}
Able, B. (1956).
\newblock Nucleic acid content of microscope.
\newblock \emph{Nature}, 135, 7--9.

\bibitem[{Able et~al.(1954)Able, Tagg, and Rush}]{AbTaRu:54}
Able, B., Tagg, R., and Rush, M. (1954).
\newblock Enzyme-catalyzed cellular transanimations.
\newblock In A.~Round (ed.), \emph{Advances in Enzymology}, volume~2, 125--247.
  Academic Press, New York, 3rd edition.

\bibitem[{Keohane(1958)}]{Keo:58}
Keohane, R. (1958).
\newblock \emph{Power and Interdependence: World Politics in Transitions}.
\newblock Little, Brown \& Co., Boston.

\bibitem[{Powers(1985)}]{Pow:85}
Powers, T. (1985).
\newblock Is there a way out?
\newblock \emph{Harpers}, 35--47.

\end{thebibliography}


\begin{thebibliography}{30}
\providecommand{\natexlab}[1]{#1}
\providecommand{\url}[1]{\texttt{#1}}
\providecommand{\urlprefix}{URL }
\expandafter\ifx\csname urlstyle\endcsname\relax
  \providecommand{\doi}[1]{doi:\discretionary{}{}{}#1}\else
  \providecommand{\doi}{doi:\discretionary{}{}{}\begingroup
  \urlstyle{rm}\Url}\fi

\bibitem[{Alonso-Mora et~al.(2018)Alonso-Mora, Beardsley, and
  Siegwart}]{mora2018cooperative}
Alonso-Mora, J., Beardsley, P., and Siegwart, R. (2018).
\newblock Cooperative collision avoidance for nonholonomic robots.
\newblock \emph{IEEE Transactions on Robotics}, 34(2), 404--420.

\bibitem[{Alonso-Mora et~al.(2015)Alonso-Mora, Naegeli, Siegwart, and
  Beardsley}]{aerialalonsomora}
Alonso-Mora, J., Naegeli, T., Siegwart, R., and Beardsley, P. (2015).
\newblock Collision avoidance for aerial vehicles in multi-agent scenarios.
\newblock \emph{Autonomous Robots}, 39(1), 101--121.

\bibitem[{Altch{\'e} et~al.(2017)Altch{\'e}, Polack, and
  de~La~Fortelle}]{highspeed}
Altch{\'e}, F., Polack, P., and de~La~Fortelle, A. (2017).
\newblock High-speed trajectory planning for autonomous vehicles using a simple
  dynamic model.
\newblock In \emph{Intelligent Transportation Systems (ITSC), 2017 IEEE 20th
  International Conference on}, 1--7. IEEE.

\bibitem[{Augugliaro et~al.(2012)Augugliaro, Schoellig, and
  D'Andrea}]{agualiro2012scp}
Augugliaro, F., Schoellig, A.P., and D'Andrea, R. (2012).
\newblock Generation of collision-free trajectories for a quadrocopter fleet: A
  sequential convex programming approach.
\newblock In \emph{Intelligent Robots and Systems (IROS), 2012 IEEE/RSJ
  International Conference on}, 1917--1922. IEEE.

\bibitem[{Bareiss and van~den Berg(2015)}]{bareiss2017general}
Bareiss, D. and van~den Berg, J. (2015).
\newblock Generalized reciprocal collision avoidance.
\newblock \emph{The International Journal of Robotics Research}, 34(12),
  1501--1514.

\bibitem[{Bekris et~al.(2012)Bekris, Grady, Moll, and Kavraki}]{Bekris2017Safe}
Bekris, K.E., Grady, D.K., Moll, M., and Kavraki, L.E. (2012).
\newblock Safe distributed motion coordination for second-order systems with
  different planning cycles.
\newblock \emph{The International Journal of Robotics Research}, 31(2),
  129--150.

\bibitem[{Bertsekas(2005)}]{dynamicprogramming}
Bertsekas, D.P. (2005).
\newblock \emph{Dynamic programming and optimal control}, volume~1.
\newblock Athena scientific Belmont, MA.

\bibitem[{Chen et~al.(2015)Chen, Cutler, and How}]{chen2015scp}
Chen, Y., Cutler, M., and How, J.P. (2015).
\newblock Decoupled multiagent path planning via incremental sequential convex
  programming.
\newblock In \emph{Robotics and Automation (ICRA), 2015 IEEE International
  Conference on}, 5954--5961. IEEE.

\bibitem[{Cheng et~al.(2017)Cheng, Zhu, Liu, Xu, and Lin}]{mpcorca}
Cheng, H., Zhu, Q., Liu, Z., Xu, T., and Lin, L. (2017).
\newblock Decentralized navigation of multiple agents based on orca and model
  predictive control.
\newblock In \emph{Intelligent Robots and Systems (IROS), 2017 IEEE/RSJ
  International Conference on}, 3446--3451. IEEE.

\bibitem[{Furrer et~al.(2016)Furrer, Burri, Achtelik, and Siegwart}]{rotors}
Furrer, F., Burri, M., Achtelik, M., and Siegwart, R. (2016).
\newblock Rotors—a modular gazebo mav simulator framework.
\newblock In \emph{Robot Operating System (ROS)}, 595--625. Springer.

\bibitem[{H{\"o}nig et~al.(2018)H{\"o}nig, Preiss, Kumar, Sukhatme, and
  Ayanian}]{honig2018}
H{\"o}nig, W., Preiss, J.A., Kumar, T.S., Sukhatme, G.S., and Ayanian, N.
  (2018).
\newblock Trajectory planning for quadrotor swarms.
\newblock \emph{IEEE Transactions on Robotics}, 34(4), 856--869.

\bibitem[{Horn(1985)}]{horn1985cr}
Horn, R.A. (1985).
\newblock Cr johnson matrix analysis.

\bibitem[{Johnson(2014)}]{nlopt}
Johnson, S.G. (2014).
\newblock The nlopt nonlinear-optimization package.

\bibitem[{Kamel et~al.(2017)Kamel, Alonso-Mora, Siegwart, and Nieto}]{rob}
Kamel, M., Alonso-Mora, J., Siegwart, R., and Nieto, J. (2017).
\newblock Robust collision avoidance for multiple micro aerial vehicles using
  nonlinear model predictive control.
\newblock In \emph{Intelligent Robots and Systems (IROS), 2017 IEEE/RSJ
  International Conference on}, 236--243. IEEE.

\bibitem[{Kraft(1988)}]{sqp}
Kraft, D. (1988).
\newblock A software package for sequential quadratic programming.
\newblock \emph{Forschungsbericht- Deutsche Forschungs- und Versuchsanstalt fur
  Luft- und Raumfahrt}.

\bibitem[{Lee et~al.(2010)Lee, Leoky, and McClamroch}]{taecontroller}
Lee, T., Leoky, M., and McClamroch, N.H. (2010).
\newblock Geometric tracking control of a quadrotor uav on se (3).
\newblock In \emph{Decision and Control (CDC), 2010 49th IEEE Conference on},
  5420--5425. IEEE.

\bibitem[{Liu and Narayanan(2011)}]{fan}
Liu, F. and Narayanan, A. (2011).
\newblock Real time replanning based on a* for collision avoidance in
  multi-robot systems.
\newblock In \emph{Ubiquitous Robots and Ambient Intelligence (URAI), 2011 8th
  International Conference on}, 473--479. IEEE.

\bibitem[{Mellinger and Kumar(2011)}]{mellinger2011minimum}
Mellinger, D. and Kumar, V. (2011).
\newblock Minimum snap trajectory generation and control for quadrotors.
\newblock In \emph{Robotics and Automation (ICRA), 2011 IEEE International
  Conference on}, 2520--2525. IEEE.

\bibitem[{Morgan et~al.(2014)Morgan, Chung, and Hadaegh}]{scpmpc}
Morgan, D., Chung, S.J., and Hadaegh, F.Y. (2014).
\newblock Model predictive control of swarms of spacecraft using sequential
  convex programming.
\newblock \emph{Journal of Guidance, Control, and Dynamics}, 37(6), 1725--1740.

\bibitem[{Mueller et~al.(2015)Mueller, Hehn, and D'Andrea}]{efficient2015quad}
Mueller, M.W., Hehn, M., and D'Andrea, R. (2015).
\newblock A computationally efficient motion primitive for quadrocopter
  trajectory generation.
\newblock \emph{IEEE Transactions on Robotics}, 31(6), 1294--1310.

\bibitem[{Rufli et~al.(2013)Rufli, Alonso-Mora, and
  Siegwart}]{rufli2013reciprocal}
Rufli, M., Alonso-Mora, J., and Siegwart, R. (2013).
\newblock Reciprocal collision avoidance with motion continuity constraints.
\newblock \emph{IEEE Transactions on Robotics}, 29(4), 899--912.

\bibitem[{Ryu and Agrawal(2011)}]{chul2010}
Ryu, J.C. and Agrawal, S.K. (2011).
\newblock Differential flatness-based robust control of mobile robots in the
  presence of slip.
\newblock \emph{The International Journal of Robotics Research}, 30(4),
  463--475.

\bibitem[{Snape et~al.(2010)Snape, Van Den~Berg, Guy, and
  Manocha}]{snape2010smooth}
Snape, J., Van Den~Berg, J., Guy, S.J., and Manocha, D. (2010).
\newblock Smooth and collision-free navigation for multiple robots under
  differential-drive constraints.
\newblock In \emph{Intelligent Robots and Systems (IROS), 2010 IEEE/RSJ
  International Conference on}, 4584--4589. IEEE.

\bibitem[{Solovey et~al.(2016)Solovey, Salzman, and
  Halperin}]{Solovey2016finding}
Solovey, K., Salzman, O., and Halperin, D. (2016).
\newblock Finding a needle in an exponential haystack: Discrete rrt for
  exploration of implicit roadmaps in multi-robot motion planning.
\newblock \emph{The International Journal of Robotics Research}, 35(5),
  501--513.

\bibitem[{Tang et~al.(2018)Tang, Thomas, and Kumar}]{tang2018hold}
Tang, S., Thomas, J., and Kumar, V. (2018).
\newblock Hold or take optimal plan (hoop): A quadratic programming approach to
  multi-robot trajectory generation.
\newblock \emph{The International Journal of Robotics Research},
  0278364917741532.

\bibitem[{Usenko et~al.(2017)Usenko, von Stumberg, Pangercic, and
  Cremers}]{2017real}
Usenko, V., von Stumberg, L., Pangercic, A., and Cremers, D. (2017).
\newblock Real-time trajectory replanning for mavs using uniform b-splines and
  a 3d circular buffer.
\newblock In \emph{Intelligent Robots and Systems (IROS), 2017 IEEE/RSJ
  International Conference on}, 215--222. IEEE.

\bibitem[{Van Den~Berg et~al.(2011)Van Den~Berg, Guy, Lin, and
  Manocha}]{Jur2011}
Van Den~Berg, J., Guy, S.J., Lin, M., and Manocha, D. (2011).
\newblock Reciprocal n-body collision avoidance.
\newblock In \emph{Robotics research}, 3--19. Springer.

\bibitem[{Van~den Berg et~al.(2008)Van~den Berg, Lin, and Manocha}]{Berg2008}
Van~den Berg, J., Lin, M., and Manocha, D. (2008).
\newblock Reciprocal velocity obstacles for real-time multi-agent navigation.
\newblock In \emph{Robotics and Automation, 2008. ICRA 2008. IEEE International
  Conference on}, 1928--1935. IEEE.

\bibitem[{Wang et~al.(2017)Wang, Ames, and Egerstedt}]{safe2017certificate}
Wang, L., Ames, A.D., and Egerstedt, M. (2017).
\newblock Safe certificate-based maneuvers for teams of quadrotors using
  differential flatness.
\newblock In \emph{Robotics and Automation (ICRA), 2015 IEEE International
  Conference on}, 3293--3298. IEEE.

\bibitem[{Zhou et~al.(2017)Zhou, Wang, Bandyopadhyay, and Schwager}]{2017fast}
Zhou, D., Wang, Z., Bandyopadhyay, S., and Schwager, M. (2017).
\newblock Fast, on-line collision avoidance for dynamic vehicles using buffered
  voronoi cells.
\newblock \emph{IEEE Robotics and Automation Letters}, 2(2), 1047--1054.

\end{thebibliography}

\section*{Appendix}
\subsection*{Partially defined end state}
In case of end position only being available, substituting the velocity and acceleration terms from costate equation Eq.\eqref{costatesol} and equating them to zero allows:
\begin{equation}
\begin{bmatrix}
T^5 & T^4&T^3 \\
-2T & -2 & 0 \\
-T^2 & -2T & -2
\end{bmatrix} \begin{bmatrix}
\beta_1 \\
\beta_2 \\
\beta_3
\end{bmatrix} = 
\begin{bmatrix}
p_{end} -(p_0 + v_0T + 0.5a_0T^2) \\
0\\
0
\end{bmatrix}
\end{equation}

The solution to the linear system results in:

\begin{equation}
\begin{bmatrix}
\beta_1 \\
\beta_2 \\
\beta_3 \\
\end{bmatrix} = \frac{1}{T^5}\begin{bmatrix}
20 \\
-20T \\
10T^2
\end{bmatrix}(p_{end}-(p_0+v_0T+ \frac{a_0T^2}{a}))
\label{equationapend}
\end{equation}

\end{document}